# AMIL: Adversarial Multi Instance Learning for Human Pose Estimation

POURYA SHAMSOLMOALI, MASOUMEH ZAREAPOOR, Shanghai Jiao Tong University, China
HUIYU ZHOU, University of Leicester, United Kingdom
JIE YANG, Shanghai Jiao Tong University, China

Human pose estimation has an important impact on a wide range of applications from human-computer interface to surveillance and content-based video retrieval. For human pose estimation, joint obstructions and overlapping upon human bodies result in departed pose estimation. To address these problems, by integrating priors of the structure of human bodies, we present a novel structure-aware network to discreetly consider such priors during the training of the network. Typically, learning such constraints is a challenging task. Instead, we propose generative adversarial networks as our learning model in which we design two residual multiple instance learning (MIL) models with the identical architecture, one is used as the generator and the other one is used as the discriminator. The discriminator task is to distinguish the actual poses from the fake ones. If the pose generator generates the results that the discriminator is not able to distinguish from the real ones, the model has successfully learnt the priors. In the proposed model, the discriminator differentiates the ground-truth heatmaps from the generated ones, and later the adversarial loss back-propagates to the generator. Such procedure assists the generator to learn reasonable body configurations and is proved to be advantageous to improve the pose estimation accuracy. Meanwhile, we propose a novel function for MIL. It is an adjustable structure for both instance selection and modeling to appropriately pass the information between instances in a single bag. In the proposed residual MIL neural network, the pooling action adequately updates the instance contribution to its bag. The proposed adversarial residual multi-instance neural network that is based on pooling has been validated on two datasets for the human pose estimation task and successfully outperforms the other state-of-arts models. The code will be made available on https://github.com/pshams55/AMIL.

This research is partly supported by NSFC, China (No: 61876107，U1803261）and 973 Plan，China (No. 2015CB856004).
H. Zhou was supported by UK EPSRC under Grant EP/N011074/1, Royal Society-Newton Advanced Fellowship under Grant NA160342, and European Union's Horizon 2020 research and innovation program under the Marie-Sklodowska-Curie grant agreement No 720325.
Authors' addresses: P. Shamsolmoali, M. Zareapoor, J. Yang (corresponding author), Institute of Image Processing and Pattern Recognition, Shanghai Jiao Tong University, Shanghai, China, emails: {pshams, mzarea, jieyang}@sjtu.edu.cn; H. Zhou, Department of Informatics, University of Leicester, Leicester LE1 7RH, United Kingdom, email: hz143@leicester.ac.uk.

# 1 INTRODUCTION

Estimation of human pose from an image is a challenging task because of the information limitation of images and the large distinctions in the form of different parts of body. Previously, most of the works used graphical models to handle these problems [1, 2, 3]. Regardless of the progresses made by mentioned fascinating models and algorithms, the bottleneck seems to be the absence of operative feature representations that have the ability to characterize several stages of visual signs and accounting for the changes in the appearance of people. Most of the recent studies [4-10] from the advanced illustration of the human body (e.g. skeleton) moved to the low-level feature collection (e.g. local features) as the full-body pose estimation is remained an effortful task. Recently, deep learning widely attracts computer vision researchers. Deep neural networks have the skills to appropriately learn better feature representations. For example, a recent proposed model reported in [11] achieved the state-of-the-art performance for human pose estimation. The distinct style, that uses repeated top-down and bottom-up inference going through different scales for different accessible fields, support the model to capture inherent relationships between human body parts. Though, such method may estimate human pose with improbable outlines because of severe occlusion or overlapping with the other neighboring people. This model predicts some similar features from the other person or the background. Nonetheless, it is challenging to integrate the priors of human body structures into Deep Convolution Neural Networks (DCNNs), as Tompson et al. [12] mentioned that the low-level DCNNs process is usually difficult to implement, whilst DCNNs have the ability of feature learning. As a result, an irrational human pose can be formed by an ordinary DCNN. As stated in [13], in case of dense occlusions, ordinary DCNNs achieve poor results. To cope with this issue, priors regarding the combination of the human body joints required to be considered. The best way to handle this problem is to learn the real body joints' structures from a huge amount of training data. Although, learning from such a distribution is a challenging task. To solve these problems, we intend to learn the distribution of human body structures. Suppose there is a discriminator which can determine the best form based on the reasonability of the predicted poses. If the model gets properly trained and generates the samples that are quite similar to the real samples and the discriminator could not distinguish the real samples from the fake ones, the model would have effectively learned the structure of the human body. In [58] the authors propose a biologically inspired appearance model for robust visual tracking. Motivated in part by the success of the hierarchical organization of the primary visual cortex, they build an architecture containing five layers: whitening, rectification, normalization, coding and polling. Zhang et al. [59] propose a novel visual model based matching framework for robust tracking based on basis matching rather than point matching. In [60] the authors present a machine learning model to learn a codebook of visual elements for representing the leaf shape and venation patterns. Zhang et al. [61] propose a real-time visual tracking method based on structurally random projection and weighted least squares techniques. Chen et al. [63] proposed a hybrid model for human pose estimation. On one hand, used feature pyramid network which can localize the "simple" keypoints like eyes and hands but may fail to precisely recognize the occluded or invisible keypoints. Then again, used RefineNet to explicitly handle the "hard" keypoints by integrating all levels of feature representations from the GlobalNet with the hard keypoint mining loss.

Due to the recent success of Generative Adversarial Networks (GAN) on several applications [14-17], we propose a discriminator to take the responsibility of checking various structures of the human body. The generator is the main human pose estimator to capture important features of the image. In the proposed model, our discriminator and generator have the same architecture. In this paper, the adversarial training approach is used to empower the discriminator to differentiate improbable poses and guide the generator. Once the training is completed, the generator can be used as a human pose estimator and the discriminator can be ignored. At the present, convolution neural networks (CNNs) are the most effective deep learning algorithms in GAN for human pose estimation [13, 17, 18, 64]. Chen et al. [65] proposed a statistical GAN based on the human biological structure. In [66] the authors presented a joint mining method based on GAN, which consists of two stacked hourglasses with a similar architecture. The typical CNNs architecture is a stack of convolutional, pooling, non-linear and fully connected layers, accompanied by a loss function [19, 20]. It is built for taking the advantages of pooling, connections, shared weights and the use of different layers to learn high-level representations of natural images, and it has shown significant performance over numerous benchmark pose estimation datasets [20, 21]. Despite that, DCNNs demand large proper labeled training data to reach superior results, however the labeling work is actually sluggish and costly by hand and if the amount of the training data is limited or with inferior quality, it will lead to suboptimal models. To minimize the influence of noisy training pairs, we model CNNs in a weakly supervised learning framework. In place of assigning labels to all the generated images, we serve the generated images as a bag and treat the main label as the bag label. This is called Multiple Instance Learning (MIL) [22, 23]. For binary MIL, a bag is labeled positive if the bag holds as a minimum positive instance, and it is labeled negative if all its instances are negative. Consequently, integrating MIL into a deep learning algorithm would fully exploit the training set's potentiality and attain better performance. Figure1 shows the performance of the proposed model on some samples.

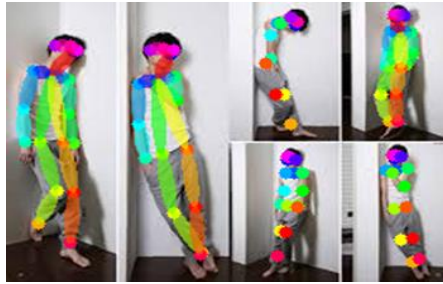

Fig. 1. The pose estimation and joint detection of the proposed model on several poses. The head and neck are indicated by purple and red respectively. The blue and green lines are specified on the right side. The light green, yellow and orange have shown on the left side.

The major contributions of this paper are three folds.

- We design a residual MIL using neural networks based on pooling to learn the configuration and structure of different human body parts through adversarial training (AMIL). The training procedure of generative adversarial networks is used to train the proposed system to solve the complex human pose estimation problems.
- To our best knowledge, we are the first one to use MIL and GAN for improving human pose estimation. We also proposed a multi-task network for predicting the pose heatmaps to achieve better performance.
- Being evaluated on two human pose estimation datasets (MPII and LSP) [24, 25], the proposed model considerably outperforms the other state-of-the-art approaches, and is able to constantly generate better pose estimation compared to the other methods.

This paper is organized as follows: Section 2 presents the related work. Section 3 illustrates the detailed description of the proposed model. In Section 4, we present the experiment results, performance evaluation of the proposed model, and we describe how AMIL improves the human pose estimation on two datasets. Section 5 concludes the paper.

## 2 RELATED WORK

This paper is related to the work using heatmaps for human pose estimation based on multiple instance neural networks and adversarial networks.

### 2.1 Human Pose Estimation

Most of the current human pose estimations methods use deep neural networks to predict the key-points of the human body in the images. DeepPose [27] is one of the first deep-learning based methods for human pose estimation, expressing the pose estimation obstacle as a regression problem by a regular convolutional model. Some modern approaches aim to predict human body structures, commonly named heatmaps or support maps that describe the probabilities to detect every keypoint at different positions. The precise place of a keypoint is predicted by calculating the maximum in combination of heatmaps. In comparison with direct regression models, heatmap-based approaches have the ability to leverage the allocated properties and are appropriate for training [27]. Several CNN architectures have been designed to capture the key evidence and cues of human body parts. Ukita and Uematsu [28] proposed a weakly-supervised learning model for human pose estimation. The authors used fully-annotated images which take a pose annotation and an action label to obtain initial pose models for every action. Papandreou et al. [29] used a CNN and geometric embedding descriptors that learn to perceive specific keypoints and estimate their relative movements that help to group keypoints into individual pose cases.

In [11], the authors used a multilevel structure to sufficiently enlarge the receptive field to learn the long-range spatial dealings. Furthermore, transitional supervision is adapted to generate intermediate assessment maps and to let them be processed over each stage. Recently, some techniques focus on dealing with the multi-person pose prediction problem. The method reported in [30] predicts multiple person poses in an image. The authors used the advantages of CNNs to produce keypoints and proposed integer linear programming to match the joints for each person in a group. In [31], the authors present a model that estimates the multiple person keypoint heatmaps and the affinity fields, and uses a greedy algorithm to assemble the joints that fit to the same person.

### 2.2 Multiple Instance learning

Multiple Instance Learning (MIL) is a different method for supervised learning. In MIL, in place of using positive or negative singletons, samples are collected to form a "bag", and each bag can have several instances [32]. Present works [23, 33, 34] indicate that, MIL delivers higher human action recognition accuracy. Ronchi and Perona [35] proposed a method to analyze the impact of errors in the algorithms for multi-instance pose estimation. The model calculates the sensitivity of a human pose with respect to the instance size, number and form of visible keypoints, mash of the instances, and the affiliate score of instances. However, the proposed model does not have significant performance in multiple human pose estimation in an image. Babenko et al. [36] showed the high performance and stability of MIL models on visual tracking and object detection. Yun et al. [37] proposed a geometric relational feature based on the distances between all the pairs of joints. They applied a method associated with MIL in which the sequence is represented by a bag of body-pose features. This model is accurate in detecting body joints but not efficient. Pathak et al. [38] proposed a new MIL model for semantic segmentation learning by using a fully convolution neural network and multi-class MIL loss. Hoffman et al. [39] provide a new formulation of a joint MIL method that contains samples from the object data and the labels, and executes field transfer learning to increase the underlying detector representation. The proposed model is efficient; however, in complex cases, the model captures a small portion of the image background as part of the objects.

Xu [40] proposed a MIL based decision neural network that attempts to bond the semantic gap in content-based image retrieval. In this model, the locally unsupervised learning in each subnet is to find the hidden structure in the "unlabeled" training data. While the negative or positive labels given to each image in the MIL-based application could be denoted as a kind of supervised information that considerably affects the training results. Zeng and Ji [41] proposed a deep CNN model, called as "multi-instance multi-task CNN", while the number of images, representing a multi-task problem, is considered as the inputs and a collective sub-CNN is linked to all the input images to build the instance representations. These models demonstrate the potential of CNN to capture the ambiguous multiple-to-multiple relationship in multi-instance multi-task learning on a data set with a limited number of the labeled samples. However, such ability for individual conformation cannot be differentiated in the experiment. Zhang et al. [42] proposed a different framework for object detection, by reformulating it as a MIL problem and additionally integrate it into a self-paced learning system. The proposed model is able to provide insightful metric measurements and learns patterns under co-salient areas in a self-learning way by using MIL. The proposed model shows that the bag level demonstration from the hierarchical compound neural network layers generate more dissimilar features than those formed by basically combing instance level representations.

### 2.3 Generative Adversarial Networks

Goodfellow et al. [43] introduced GANs for generating natural images such that it allows for unsupervised training while minimizing the blur consequence of using variational autoencoders. However, there are some concerns about hard training and instability of GANs. To efficiently train GANs, several models have been proposed to use deep convolutional architectures [18]. These models presented some elements in their networks to improve the stability, for instance, using batch normalization to avoid diversity loss and removing the fully connected layer (FC). The combination of deep convolution and GAN leads to an impressive configuration to handle the training of GAN. Gulrajani et al. [20] proposed to use gradient penalty instead of the weight clipping strategy. Gradient penalty is a loss function and an extra term that adapts the discriminator's gradient norm.

Table 1. Summary of some key achievements for pose estimation.

| Ref. | Approach | Experimental Results on MPII | Optimization | Efficiency |
|---|---|---|---|---|
| [6] | CNN and heatmap labeling | PCKh = 88.1 | per-pixel Softmax loss | --- |
| [11] | Hourglass CNN | PCKh = 90.9 | Rmsprop | --- |
| [12] | CNN and body joints relations | PCKh = 90.2 | Coarse optimization | Input Convolution stages |
| [13] | GAN and structure-aware CNN | PCKh = 92.1 | Confidence discriminator | --- |
| [17] | GAN and stacked hourglass CNN | PCKh = 91.8 | per-pixel loss | --- |
| [26] | Deep CNN | PCKh = 90.8 | Optimal Back-propagation | --- |
| [27] | Dense correspondences between image and a human body | PCKh = 91.7 | --- | Annotation pipeline |
| [35] | MIL algorithm error optimization | --- | Optimal MIL | --- |
| [37] | MIL geometric relation | PCKh = 91.1 | --- | --- |
| [47] | CNN and multi context attention | PCKh = 91.5 | RMSprop algorithm | Generate attention maps |
| Ours | Adversarial MIL and heatmap labeling | PCKh = 92.3 | Gradient descent and Pixel loss | Adjust pooling |

The overall performance proves that this strategy is more stable and faster than the other methods. Another work by Berthelot et al. [44] presents a balance term based on the relational control theory, to make equilibrium between the generator and the discriminator. Meanwhile, if the model is collapses or reaches its last state, a conversion measure is used to control the process. In [67] the authors present a dual discriminator generative adversarial network which, unlike GAN, equipped with two discriminators; and together with a generator, it also has the analogy of a min-max game, wherein a discriminator rewards high scores for samples from data distribution whilst another discriminator, conversely, favoring data from the generator, and the generator produces data to fool both two discriminators. Hoang et al. [68] proposed to use multiple generators, instead of using a single one to overcome the collapsing problems. In [69] the authors proposed to use several generator and discriminator to increase the performance of GAN. Nevertheless, the model required high computational resources.

GANs have a great success on generating images [18]; hence, it is highlighted for unsupervised learning. The idea of uncertain GAN [45] is presented for combining the class information. These methods merge the loss of uncertain GAN and the $L_1$ or $L_2$ gap between the generated and the ground-truth images. Another technique is to creating heatmaps of labels similar to semantic segmentation [46], or human pose recognition [13]. Chu et al. [47] proposed to integrate CNNs and a multi-context attention model into an end-to-end framework to recognize human poses. The model used hourglass networks to produce heatmaps from features at multiple resolutions with different semantics. Using the adversarial training approach can bring certain benefits. In this paper, we use adversarial training methods [44] to increase the performance of the pose estimator. Pose estimation can be considered as a conversion from an RGB image to a multichannel heatmap. The proposed network can well achieve this translation. Dissimilar to the other works, in the proposed network, the discrimination objective is not only to distinguish a fake image from a real one, but also to enforce the geometric constraint to the model. Table 1 summarizes some key achievements for pose estimation, where we listed each approach, experimental results, computational complexity, and efficiency.

The traditional pose estimation pipeline is based on three major points: 1) extraction of local features; 2) dictionary learning and usage of feature encoding, and 3) classification of the actions. Dense and improved trajectories [27], is popular since it highlights local changes in the spatio-temporal domain. On the other hand, local feature descriptors are also pooled to obtain image and video-level representations [18]. In addition to local interest points or feature descriptors, mid-level features of body parts [17] or deep features [54] are used frequently to discover abstract representations from videos. A detailed discussion on different video-based feature extraction techniques can be found in [56]. Nevertheless, instead of detecting local interest points, an alternative way is to extract all the local descriptors while filtering out the effects of the un-representative ones throughout the learning process. Specifically, the MIL paradigm assumes the images and videos as bags of instances (local descriptors) without concerning about their discriminative skills. However, in the MIL origination, it is required that a given bag contains at least one class precise descriptor.

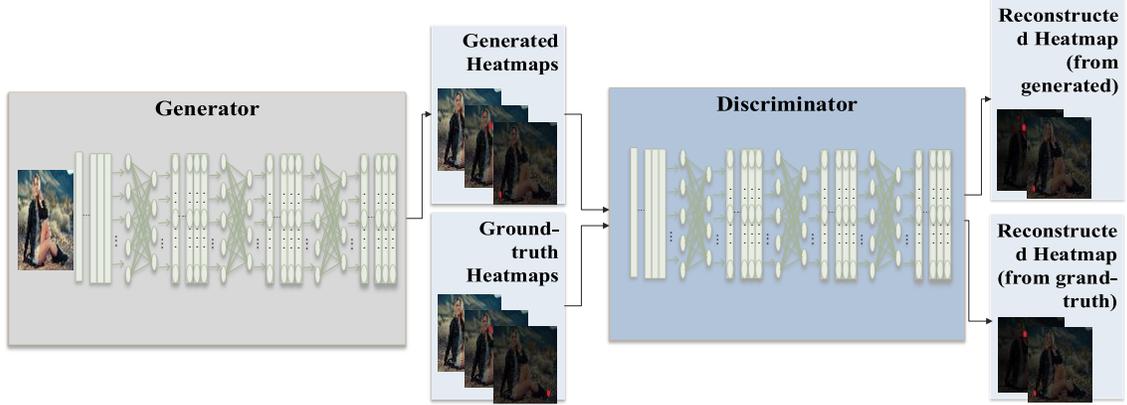

Fig. 2. The proposed model framework. We propose a combination of MI_RNet for pose estimation as the generator (on the left) with a discriminator (on the right) to differentiate the ground-truth heatmaps from the generated heatmaps by input heatmaps reconstruction.

## 3  THE PROPOSED ADVERSARIAL MULTI-INSTANCE LEARNING

As presented in Figure 2, the proposed AMIL model consists of two networks, the pose generator and the pose discriminator. The first network, generator, is a multiple instance residual neural network (MI-RNet) architecture. The inputs to the generator are the RGB images after the processing unit; it generates a set of heatmaps for each input image that specifies the confidence score for every keypoint on different locations of the image. The other network, the discriminator, has architecture similar to the generator; it encodes the heatmaps along with the RGB image and decodes them into a new set of heatmaps in order to distinguish real heatmaps from fake ones. Moreover, after training the pose generator with the guidance of the pose discriminator, the human body priors are extracted, which helps to increase the prediction accuracy.

### 3.1  Pose Generator

The generative network objective is to learn a mapping from an RGB image to keypoint heatmaps. Figure 3 illustrates the architecture of the generator. If the model extracts clear information of the body parts, it offers significant materials for describing the geometric information of a human pose. The goal of the generative network is to learn and project an image $y$ onto the corresponding pose heatmaps $x$, $G(y) = \{\hat{x}\}$ while $\hat{x}$ is the predicted heatmap. MI-RNet has the ability to learn contextual features from the input images. Furthermore, in the pose discriminator, the adversarial loss is introduced and used for presenting the error between the ground-truth heatmaps and the generated ones. This method supports the generator to learn the f spatial dependencies from the input images and the human body patterns.

The basic block of the pose generator is expressed as follows:

$$\begin{cases} \{X_n, Y\} = G_n(X_{n-1}, Y) & if\ n \geq 2 \\ \{X_n, Y\} = G_n(Y) & if\ n = 1 \end{cases}$$

$X_n$ is the output initiation tensor of the $n_{th}$ weighted generative network for pose detection. $Y$ is the image feature, captured after pre-training by using the proposed MI-RNet. In the proposed model, the final heatmap output $\hat{x}_n$ is achieved from $X_n$ through the FC layers with the step size of $1$ without padding. Specifically, the performance of the final FC layer performances as a linear classifier is gained as the final predicted heatmaps. Consequently, the specified training set $\{y^i, x^i\}_{i=1}^{M}$ and $M$ represents the number of the images that are assigned for training. Moreover, the adversarial loss for the pose discrimination is proposed and considered jointly with the errors between the ground-truth heatmaps and the generated ones. This method supports the generator to learn the features and spatial dependencies out of the images and also detect all the human body configurations.

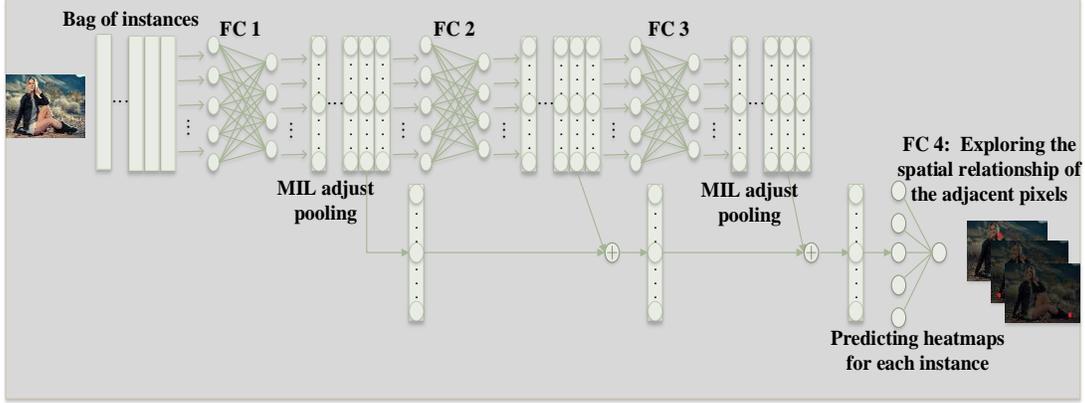

Fig. 3. The architecture of the proposed residual multiple instance neural network. The first FC layer produces a bag feature vector. The next FC layers learn the residuals of bag representation. The sizes of all the FC layers are set to 128.

### 3.1.1 Multi Instance Residual Neural Network (MI-RNet) Architecture

MIL focuses on handling the intricate data in the form that all the bags $X = \{X_1, X_2, ..., X_N\}$ and instance features of $i_{th}$ bag $X_i = \{x_{i1}, x_{i2}, ..., x_{imi}\}$, $x_{ij} \in R^{d \times 1}$, while $N$ and $m_i$ represent the number of the bags and instances in bag $X_i$ correspondingly. Assume $Y_i \in \{0, 1\}$ and $y_{ij} \in \{0, 1\}$ distinctly are the label of bag $X_i$ and instance $x_{ij}$, where 0 and 1 represent negative and positive respectively. In MIL, simply bag labels are provided throughout the training, and there are two limitations for MIL:

• In case that bag $X_i$ is negative, therefore all the instances in $X_i$ will be negative, i.e., whenever $Y_i = 0$, then all $y_{ij} = 0$;
• In case bag $X_i$ is positive, therefore at least one instance in $X_i$ will be positive, i.e., whenever $Y_i = 1$, then $\sum_{j=1}^{m_i} y_{ij} \geq 1$.

The aim of MIL is to train a bag classifier to estimate a new bag label. In the MI network, instance-to-bag connections are various under different hypotheses. Therefore, a constant MI hypothesis on instance labels and bag labels are not signified. Accordingly, we endeavored to create a MI model to predict the bag labels. Unlike the pixels having the spatial relation, in MIL, the bag instances are a set of features that do not have a precise order. Hence, a significant asset of the MI data is the invariance to input permutation. MI networks [22] have three phases: (1) learning an instance embedded by the instance modifier; (2) executing a permutation-invariant MIL pooling to create an improvised bag; (3) bag classification depending on the bag embedding. Every phase has the permutation-invariant assets plus the essential theorem of symmetric functions [48]. Deep residual learning is proposed recently [49] and impressively performed in object detection by taking advantages of deep neural networks. In this work, as represented in Figure 3, three FC layers and three proposed MIL adjusting pooling layers are implemented. For each middle FC layer that can learn instance features, a FC layer for predicting instance scores and a proposed MIL adjust pooling layer follows it. During the training, the supervision is added to each level. During the testing, we compute the mean score for each level. We apply the residual connections after each MIL adjust pooling layer to concatenate the instance features. The task of the first FC layer is to produce a bag of feature vector. The next fully connected layers learn the residuals of bag representation. The sizes of all the FC layers are set to 128. MI-RNet is formulated as:

$$\begin{cases} x_{ij}^l = H^l(x_{ij}^{l-1}), \\ X_i^1 = M^l(x_{ij|j=1...m_i}^1), \\ X_i^l = M^l(x_{ij|j=1...m_i}^l) + X^{l-1}, l > 1. \end{cases} \quad (1)$$

For example, a single bag $X_i$ have various instances $x_{ij}$, in the MI-RNet. It is composed of $l$ layers which contain a non-linear transformation $H^l(.)$, while $l$ indicates the layer. $H^l(.)$ is a compound of several actions such as rectified linear units (*ReLU*), or inner product (or FC) [50]. The output of the $l_{th}$ layer of an instance $x_{ij}$ is represented as $x_{ij}^l$. In $X_i^{l,k}$, the $k_{th}$ index means multiple bag features from all different levels of instance features that have been learned by MIL adjust pooling. MI-RNet, by utilizing multiple hierarchies, can achieve better bag classification accuracy. We could expound it in two folds: first, for instance, better feature training can be achieved at bottom layers; and second, for testing, the average of multiple bag probabilities will be calculated to catch a better label. The weights of different levels are set equally in this work. The proposed MI-RNet is not similar to the standard residual learning [49] which learns representation residuals by convolution, *ReLU* and batch normalization; the model learns the residuals of the bag representation through the FC layers, *ReLU* and MIL adjust pooling. At the last stage of the network, the final bag representation is connected to the bag label via a FC layer with one neuron and softmax activation [50].

### 3.1.2 MIL Adjust Pooling

Other MIL pooling approaches find it is difficult to set the contextual information among the instances in a bag, as the pooling functions are of feed-forward procedures and the instance weights are calculated separately. Motivated by [51, 52], we recommend using adjust pooling. To demonstrate the process, *f(.)* represents the instance transformer and $f(X) = \{f(x_1), f(x_2), ..., f(x_K)\}$ signifies the instance embedding corresponding to the bag *X*. The proposed adjust pooling can be stated as a weighted-sum pooling step:

$$\sigma(X) = \sum_{i=1}^{k} w_i f(x_i), \tag{2}$$

where $w_i$ denotes the instance weight which defines the influence of the instance $i_{th}$ onto its bag embedding. According to these weights, to combine the instance embeddings into a single bag embedding in a weighted-sum pooling model, we use a non-linear squeezed function. The non-linear squeezed function [51] is formulated as follows:

$$s(X) = \frac{||\sigma(X)||^2}{1 + ||\sigma(X)||^2} \frac{\sigma(X)}{||\sigma(X)||}. \tag{3}$$

The instance weight $w_i$ can be computed using an adjustable method. To illustrate the pooling process, a provisional instance weight is defined as $b_i$. Afterwards, the instance weight $w_i$ is appointed via a straightforward function as follows.

$$w_i = \frac{\exp(b_i)}{\sum_j \exp(b_j)}. \tag{4}$$

The superscript *t* signifies the $t_{th}$ iteration. First, *t = 1* and $b_i^1 = 0$, which shows that each instance equally contributes to embedding a single bag. Later, the instance weights are updated regularly while their similarities are considered in the last updated embedded bag. Hence, in the $t_{th}$ iteration, the embedded bag is $s^t(X)$; consequently the update function for temporary instance weight $b_i^t$ is as follows:

$$b_i^{t+1} = b_i^t + f(x_i).s^t(X). \tag{5}$$

After every feed forward pass, we can implement the bag embedding $s^T(X)$. $L_2$ norm is performed on $s^T(X)$ to demonstrate the probability of the positive bag indicated as *//s//*. Therefore, the proposed MI-RNet can be optimized in the following form while Y denotes the bag label, $m^+ = 0.9$ and $m^- = 0.1$.

$$L(X) = Ymax(0, m^+ - ||s||)^2 + (1 - Y) \max(0, ||s|| - m^-)^2. \tag{6}$$

Subject to the major proposition of symmetric functions [48, 57], the permutation-invariant symmetric functions *W* can be formulated as follows:

$$W(X) = \rho(\sum_{x \in X} \emptyset(x)). \tag{7}$$

where ρ and φ denote the transformations. To verify the permutation-invariant of the proposed adjust pooling, we demonstrate that the proposed pooling method satisfies the permutation-invariant symmetric requirements. The aim of the proposed pooling model is to employ weighted-sum pooling. In this model, the weight reflects other instances fitting to the same bag and its total value is calculated by *t* time's iteration. At the beginning (*t* = 1), adjust pooling starts with mean pooling:

$$\partial^1(X) = \sum_i w_i^1 f(x_i), \tag{8}$$

While $\forall_i \in [1, K]$ $w_i^1 = \frac{i}{K}$. The mean pooling is a classic symmetric function. In the $t_{th}$ iteration, the pooling function is calculated as:

$$\partial^t(X) = \sum_i w_i^t . f(x_i) = \sum_i softmax\left(\sum_{t>1} f(x_i)s^{t-1}(X)\right)f(x_i). \qquad (9)$$

At the $t_{th}$ iteration, the bag embedding is $s^t(X)$, which is the result of the symmetric function and retains the property of permutation-invariant. Based on the symmetric function shown in Eq. (7), we know the adjust procedure is part of $\emptyset$ and $L_2$ norm that signifies the position of $\rho$ and calculates the bag length.

### 3.1.3 Training the Generator

To use multiple instance learning for pose estimation, different regions in each image should be considered as a bag. If $B = \{x^1, x^2 ...x^m\}$, $x^m$ is the *m-th* region in the image. The loss function of the bag $B$ is calculated as follows:

$$f_{loss} = -\sum_{i=1}^{S} y_i \log(p(S_i = 1|B), \qquad (10)$$

Here $p(S_i = 1/B)$ denotes the possibility that the bag is correctly classified into the $i_{th}$ class and $y_i = \{0, 1\}^{1 \times S}$ is the label matrix, while $S$ is the total sum of the classes. Based on the MIL theory, if all the instances in the bag are negative, $B$ is a negative bag for the $i_{th}$ class:

$$p(S_i = 0|B) = \prod_{j=1}^{m}\left(1 - p(S_i = 1|X^j)\right), \qquad (11)$$

$p(S_i = 1/x^j)$ is the probability of the $j_{th}$ image region selected as the $i_{th}$ class, and

$$p(S_i = 1|X^j) = 1 - \exp(-\lambda h_i^j). \qquad (12)$$

where $h_i^j$ is the $i_{th}$ output of the proposed MI-RNet model before the loss layer for $j_{th}$ region, and $\lambda$ is a constant positive value and $h_i^j \in [0, \infty)$. Eq. (12) not only significantly improves the classification results but also simplifies the gradient's calculation. For the MI-RNet, we propose a loss function to perform in parallel with the learning of the pooling function along with the instance-level classifier. In the typical MIL, the supervision of the bag-level should be handled in the loss function. As an example, the cross entropy can be used as the bag-level loss function, where $C$ is the number of possible classes and $c \in \{1, ..., C\}$. On the other hand, due to unavailability of the instance-level labels $y_{ij}^c$, we just rely on the bag-level label $y_i^c$.

$$loss_i^c = -y_i^c \times \log(p_i^c) - (1 - y_i^c) \times \log(1 - p_i^c). \qquad (13)$$

Furthermore, as the adjust pooling function is considered for gathering the instance-level predictions to attain the bag-level prediction, the ideal parameters of pooling should be dependent to the efficiency of the instance-level, which is used during the training procedure. Eq. (14) is derived from Eq. (13) to estimate the instance-level loss

$$If_{loss} = -u_{ij}^c \times \log(q_{ij}^c) - (1 - u_{ij}^c) \times \log(1 - q_{ij}^c), \qquad (14)$$

while the relevant likelihoods to the class $c$, are denoted by $q_{ij}^c \in [0,1]$ and $u_{ij}^c = 1$ ($q_{ij}^c \geq 0.5$) also 1( ) is an pointer function. In fact, the instance-level loss is a weight of the uncertainty of $q_{ij}^c$, which furthermore signifies the discriminative skill of the instance-level classifier. Additionally, we propose to minimize the loss difference between the instance level and the bag-level as follows,

$$d_i^c = ||loss_i^c - \frac{1}{N_i}\sum_{j=1}^{N_i} If_{loss}||. \qquad (15)$$

Adjust pooling is used as a bridge between instances and bags to minimize Eq. (15), and to fit it for the current status of the instance-level classifier. The task of the pooling layer is to efficiently transfer the instance-level discriminative skill into the bag-level classification skill. While training in the perspective of error back propagation, the loss at the bag-level is back propagated to the instance-level classifier over the pooling layer. Therefore, the learned pooling is considered to be ideal for training. To wrap up, the following loss function is proposed to mutually minimize the bag-level loss and the gap between the bag-level and the instance-level losses,

$$f_{loss} = loss_i^c + \lambda(d_i^c)^2 = loss_i^c + \lambda \left(loss_i^c - \frac{1}{N_i}\sum_{j=1}^{N_i} If_{loss}\right)^2, \qquad (16)$$

while $\lambda=1$ and is the Lagrange multiplier and for the mathematical simplicity, the square of difference Eq. (15) is used. In the proposed model the gradient descent approach is used to optimize the loss function for training the deep networks. In a mini-batch, the losses of different bags and instances are considered separately and then summarized. Meanwhile, losses of negative and positive bags are not computed similarly. For the positive bags where $y_i^c = 1$, the loss function presented in Eq. (16) is used. However, for the negative bags while $y_i^c = 0$, the instance-level loss is directly adopted,

$$f_{loss} = \frac{1}{N_i}\sum_{j=1}^{N_i} If_{loss}, \qquad (17)$$

where Eq. (14) is used to calculate $If_{loss}$ and $u_{ij}^c = 0$. Therefore, the gradients are formulated as,

$$\frac{\partial f_{loss}}{\partial h_i^j} = \lambda(1 - y_i). \qquad (18)$$

The methods of dynamic pooling [51], adaptive pooling [52], dynamic routing [56], and adjust pooling are adopted in the part-to-whole connection strategy. The softmax and sigmoid functions are also used to sort out the learned weights and then perform the weighted-sum pooling.

It is worthy to mention that the task of the softmax function is different. In dynamic pooling [51], softmax is applied into the individual bags to extract the relationships among them and in the adaptive pooling [52] to handle the same task sigmoid is used. In the dynamic routing [56], the softmax is used for weighting all the parent capsules with only a single child capsules. Hence, each particular weight means that the ratio of the consistent capsule in the overhead layer is sent to the child capsule. However, in the adjust pooling; the weight shows the instance involvement to the bag embedding. The softmax function is applied to all the instance contributions of the same bag and pushes them to interact with each other.

### 3.2 Pose Discriminator

To empower the training of the model in extracting the configurations of human body joints, the pose discriminator is designed. The discriminator's task is to recognize actual images from the generated ones. The discriminator input contains the heatmaps of the ground-truth images or the generated images, which are integrated with the corresponding actual images of persons. The discriminator should learn from the input pairs that the pose demonstrated by the heatmaps is accurate and matches the human pose in the input images. Meanwhile, the other task of the discriminator is to reconstruct a new set of heatmaps. The reconstructed heatmaps, similar to the real ones, help to determine the efficiency of the discriminator. The loss is calculated as the error between the real and the reconstructed heatmaps. For every single training image, the generated and the base heatmaps will be fed to the discriminator. The batches of the heatmaps will be reconstructed to compute $l_{real}$ and $l_{fake}$. The discriminator is updated at each iteration by using the collected gradient, which is computed according to $l_{real}$ and $l_{fake}$. When the input contains the ground-truth heatmaps, the discriminator is trained to identify it and create a similar one, whilst reducing the error between the reconstructed heatmaps and the groundtruth ones. The loss is formulated as,

$$l_{real} = \sum_{j=1}^{M}(S_j - D(S_j, X))^2,$$
$$l_{fake} = \sum_{j=1}^{M}(\hat{S}_j - D(\hat{S}_j, X))^2,$$
$$l_D = l_{real} - k_i l_{fake}. \qquad (19)$$

Pixel loss ($l_D$) is used to optimize the discriminator. The discriminator for each particular pose based on the bunch of heatmaps generates a value for each pixel. This value is the discriminator's error rate. The value shows the correctness level of a particular pixel from the view of the discriminator. For instance, if the left elbow is more accurate than the right elbow, a proper trained discriminator will create a heatmap of the left elbow that has a larger error at the position of the right elbow. This is dissimilar to a conventional GAN that only judges the properness of the whole input. Here, on the input heatmaps, the discriminator offers detailed comments and advises which parts of the heatmaps do not yield a real pose. In addition to the adversarial loss, $L_2$ loss is also applied on the predicted poses to measure the difference between the generated and the actual ground truth heatmaps. The final loss is the sum of the adversarial loss and an $L_2$ loss. A variable $k_t$ is used for balance controlling between the generator and the discriminator [44]. The variable $k_t$ is defined as follows and updated at every iteration $t$.

$$k_{t+1} = k_t + \omega_k(\gamma l_{real} - l_{fake}),  \qquad (20)$$

where $k_t$ is limited between 0 and 1, and $\omega_k$ is a hyperparameter. From Eq. (19), $k_t$ shows the amount of the emphasis put on $l_{fake}$. If the generator performs better than the discriminator, $l_{fake}$ is smaller than $\gamma l_{real}$, and the generated heatmaps are very similar to the real ones. Therefore, $k_t$ will be increased to make $l_{fake}$ more dominant; and accordingly the discriminator will be trained to better recognize the generated heatmaps. Similarly, if the discriminator performs better than the generator, $k_t$ will be decreased to slow down the training on $l_{fake}$ thus the generator can keep up its performance with the discriminator.

### 3.3 Adversarial Networks Training

The training of the proposed adversarial network is based on supervised learning. The goal of the generator is to reduce the gap between $\hat{S}$ and $D(\hat{S}, X)$, however, the discriminator attempts to increase it. To differentiate different poses, the discriminator tries to detect the important aspect of the real pose distribution throughout the reconstruction process. Simultaneously, the generator tries to improve the quality of human pose heatmaps thus it may fool the discriminator and to allow the discriminator to reconstruct the same heatmaps. The discriminator can be eliminated after the completion of the training. The generated heatmaps $\hat{S} = G(X)$ will be used to conclude the last outcome. To conduct the estimations, the original image and its flipped form are evaluated, and their output heatmaps are averaged. During the training, the location with the largest confidence score in each joint's heatmap is extracted. Then, the model converts the location to the original coordinate space with respect to the input image size.

## 4 EXPERIMENTS

The proposed model is evaluated on the two benchmark pose estimation datasets, MPII Human Pose [1] and extended Leeds Sports Poses (LSP) [3]. The MPII dataset contains 25,000 images with 40,000 annotated samples, around 28,000 for training, and 11,000 for testing. The whole body images are annotated with 16 different landmarks and several directions to the camera. The images are taken from videos in YouTube and the contents include daily human activities. Compared to the other human pose datasets, MPII has affluent information, for example, fully unannotated image frames. During the training, only keypoint positions are used. The LSP dataset contains 11,000 training images and 1,000 testing images that show different sports activities. The performance of the proposed model is demonstrated using the UCF YouTube action dataset [55]. The proposed network is composed of three fully connected layers that all have a size of 128 and adjust pooling function. The weights of the fully connected layers are initialized by a normal distribution and biases are initialized to 0. $T$ is assigned to 3 which denotes the iteration times of the adjust pooling. The Adam optimizer is used to optimize the network [53].

The details of the hyper-parameter optimization process such as learning rate and weight decay are illustrated in Table 2. For the MPII and LSP datasets, after every 20 iterations, the learning rates are decayed with the base 0.01 and 0.005 respectively. The provided hyper-parameters are specified by the model selection system based on the highest validation performance. Five times of 10-fold cross validation independently are run and we use the average results as the final results. The proposed model is implemented in TensorFlow 1.3.0 GPU as the backend deep learning engine. Python 3.6 is used for all the implementations. All the implementations of the network are conducted on a workstation equipped with an Intel i7-6850K CPU with 64 GB Ram and an NVIDIA GTX Geforce 1080 Ti GPU and the operating system is Ubuntu 16.04.

Table 2: The optimization procedure of the hyper-parameters

| Dataset | Learning rate | Weight decay | Iterations | Decay steps |
|---------|---------------|--------------|------------|-------------|
| MPII    | 0.001         | 0.01         | 350        | 20          |
| LSP     | 0.001         | 0.005        | 350        | 20          |

## 4.1 Evaluation Metrics

The experiments are based on two metrics. PCK is used to measure the performance on LSP. For MPII, PCKh is used.

Percentage of Correct Keypoints (PCK) [54]:
PCK shows the percentage of the precise detection that is located within a tolerance range. The tolerance range is a portion of the torso size. It can be formulated as,

$$\frac{||y_i - \hat{y}_i||_2}{||y_{lhip} - y_{rsho}||_2} \leq r, \quad (21)$$

while $y_i$ is the ground-truth location of the $i_{th}$ keypoint and $\hat{y}_i$ is the estimated location of the $i_{th}$ keypoint. The fraction $r$ is limited between 0 and 1.

Percentage of Correct Keypoints with respect to head (PCKh) [1]:
PCKh is very similar to the PCK, where the tolerance range is a fraction.

## 4.2 Results and Evaluations

Table 3 reports the performance of the proposed AMIL model and the other approaches on the LSP dataset based on PCK. The results of the proposed AMIL model is shown in Table 4 with the MPII training set, and the results are computed at $r = 0.2$. AMIL has the best detection rate through all the tolerance range. Moreover, at the tighter distance ($0.05 < r < 0.1$), the proposed model demonstrates much better outcomes.

Table 3. Human pose detection on the LSP dataset based on PCK.

| Methods | Head | Sho. | Elb. | Wri. | Hip | Knee | Ank. | Mean |
|---------|------|------|------|------|-----|------|------|------|
| [12]    | 94.8 | 88.7 | 81.3 | 76.8 | 83.6 | 86.7 | 81.9 | 84.7 |
| [25]    | 94.9 | 88.7 | 81.5 | 76.9 | 83.5 | 86.9 | 82.3 | 84.9 |
| [4]     | 95.2 | 89.0 | 81.5 | 77.0 | 83.7 | 87.0 | 82.8 | 85.2 |
| [30]    | 96.8 | 95.2 | 89.3 | 84.4 | 88.4 | 83.4 | 78.0 | 88.6 |
| [6]     | 97.1 | 92.1 | 88.1 | 85.1 | 92.2 | 91.5 | 88.7 | 90.7 |
| [11]    | 97.0 | 92.3 | 88.2 | 85.2 | 92.2 | 91.6 | 88.9 | 90.8 |
| [47]    | 98.1 | 93.7 | 89.3 | 86.9 | 93.4 | 94.0 | 92.5 | 92.6 |
| [62]    | 98.4 | 93.8 | 89.7 | 87.4 | 93.9 | 94.0 | 92.8 | 92.9 |
| [13]    | **98.5** | 94.0 | 89.8 | 87.5 | 93.9 | 94.1 | 93.0 | 93.1 |
| [17]    | 98.2 | 94.9 | **92.2** | 89.5 | **94.2** | 95.0 | 94.1 | 94.0 |
| Ours    | 98.4 | **95.1** | **92.2** | **89.7** | 94.0 | **95.3** | **94.2** | **94.2** |

Figure 4 presents exemplar qualitative performance of the proposed model. We can see that our proposed model achieves better understanding which leads to correct human body joint detection and pose estimation. In Figure 4, there are a range of images in different poses (highly twisted, partly occluded, complex structures in the wild and invisible body limbs) that our proposed AMIL successfully predicts the joints and estimate the poses. This is due to the shape prior learning and correct extraction of the proper features in the training process. However, some state-of-art models locate some of the body parts to the wrong place due to the absence of the correct body configuration constraints. The proposed discriminator contains the body constraints; hence the network successfully pinpoints the precise body position even in some challenging situations.

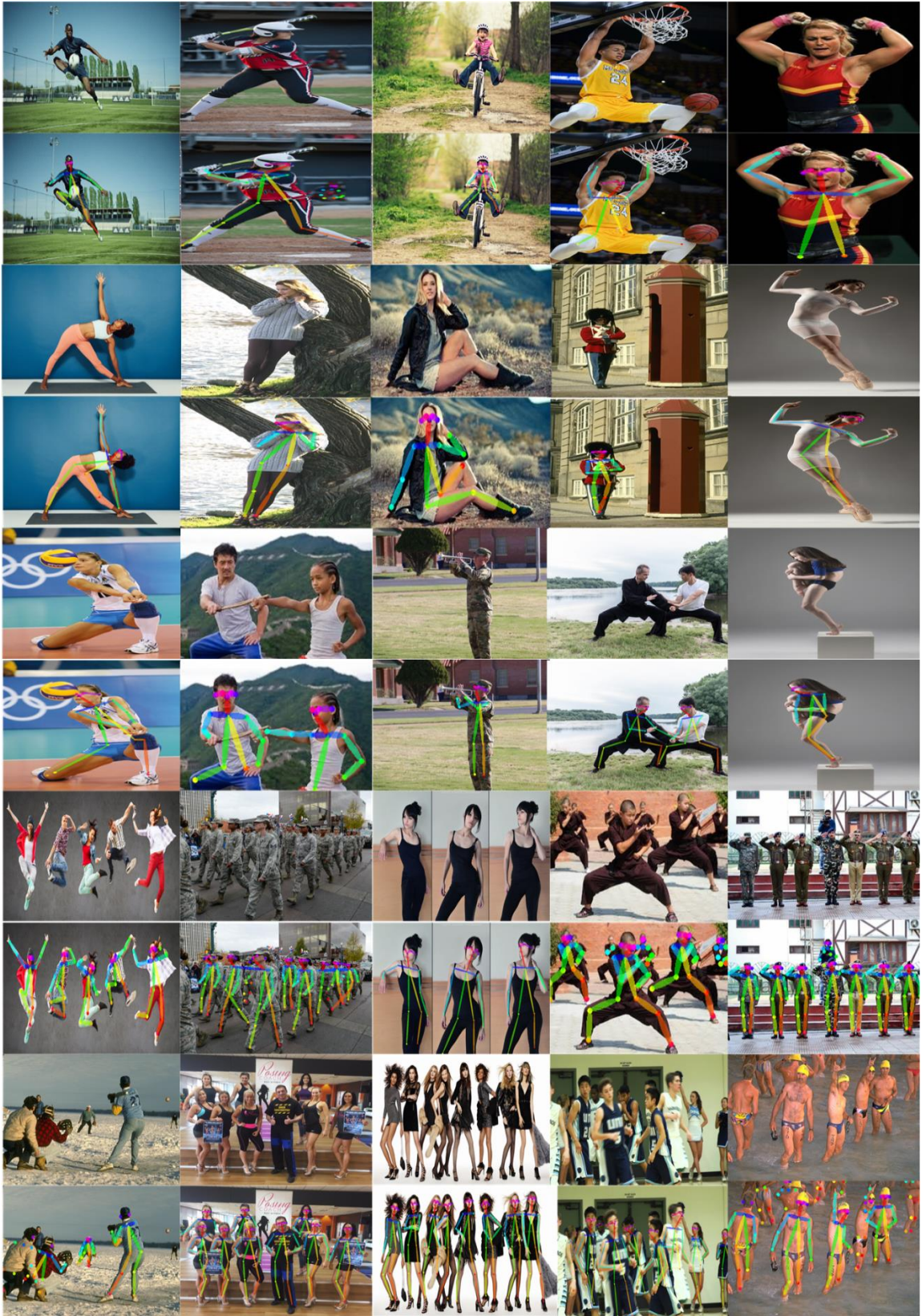

Fig. 4. Qualitative results of AMIL. The blue lines indicate the right arm and green lines indicate the body right side, the Caribbean green lines indicate the left arm and yellow and orange lines indicate the left side of body.

Table 4. Human pose detection on the MPII dataset based on PCKh.

| Methods | Head | Sho. | Elb. | Wri. | Hip | Knee | Ank. | Mean |
|---|---|---|---|---|---|---|---|---|
| [12] | 95.8 | 90.3 | 80.5 | 74.3 | 77.6 | 69.7 | 62.8 | 79.6 |
| [25] | 96.1 | 91.9 | 83.9 | 77.8 | 80.9 | 72.3 | 64.8 | 82.0 |
| [26] | 96.1 | 92.0 | 84.1 | 77.9 | 81.1 | 72.3 | 64.9 | 82.1 |
| [4] | 97.7 | 95.0 | 88.2 | 82.9 | 87.9 | 82.6 | 78.4 | 88.2 |
| [30] | 96.9 | 95.3 | 89.4 | 84.5 | 88.5 | 83.5 | 77.9 | 88.6 |
| [6] | 97.9 | 95.1 | 89.9 | 85.4 | 89.4 | 85.6 | 81.8 | 89.7 |
| [11] | 98.2 | 96.2 | 91.2 | 87.1 | 90.2 | 87.5 | 83.6 | 90.9 |
| [47] | 98.5 | 96.3 | 91.9 | 88.1 | 90.6 | 88.0 | 85.0 | 91.5 |
| [62] | 98.3 | 96.5 | 92.1 | 88.0 | 91.1 | 88.9 | 85 | 91.6 |
| [17] | 98.2 | **96.8** | 92.2 | 88.0 | 91.3 | 89.1 | 84.9 | 91.8 |
| [13] | 98.6 | 96.4 | 92.4 | **88.6** | 91.5 | 88.6 | 85.7 | 92.1 |
| Ours | **98.8** | 96.5 | **92.5** | 88.5 | **91.5** | **88.8** | **85.8** | **92.3** |

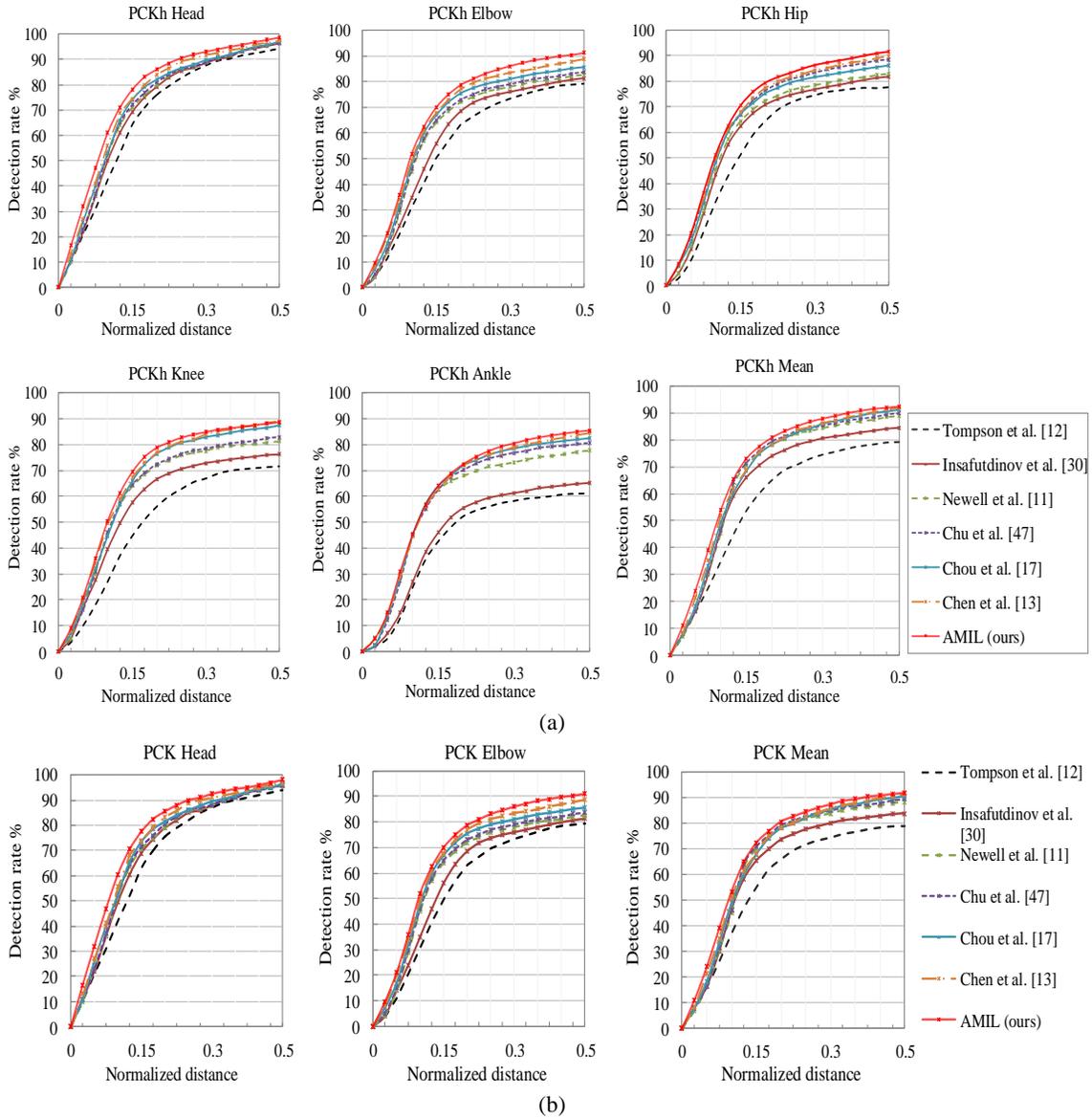

Fig. 5. (a) PCKh on MPII dataset, and (b) PCK on the LSP dataset.

Performance of AMIL and previous methods at *r = 0.5* on the MPII dataset is presented in Table 4 and Figure 5 (a). AMIL is trained with the LSP training set. Figure 5 (b) shows the performance of AMIL in comparison with other models on LSP dataset.

### 4.3 Analysis

Here, we illustrate the influence of different factors on AMIL. The experiments have been conducted on the MPII test dataset and the accuracy through training iterations has been recorded.

#### 4.3.1 MIL and MI-RNet

To evaluate the performance of the proposed AMIL model, we have conducted experiments on several network configurations. In Figure 6 (a), we compare the prediction of the standard MIL [32] and Residual MIL on the MPII dataset. Figure 6 (b) presents the performance of MI_RNet with and without adjust pooling. As the result shows, by using residual MIL and adjust pooling, there are significant improvements in the pose estimation. The discriminator shows satisfactory performance even while the person image is not provided. The reason is that the pose even could be estimated by only the pre-trained pose information. The image of the person is additional information; however the discriminator does not require this information all the time. Meanwhile, we compare the result of the MI-RNet generator trained with that of the discriminator while applying adjusts pooling with the standard MIL. These networks are trained by using the heatmaps. The performance of the proposed body-structure-aware adopted GANs on the MPII validation set increases by 0.7% compared to the standard model. This result proves that the discriminator guides the generator to generate more reliable poses that look similar to the ground truth heatmaps. In fact, separately adding the MI-RNet with adjust pooling or discriminator increases the pose estimation accuracy. However, adopting them separately improved the results by 11.2% and 0.6% respectively, though by both design adoption makes the overall improvement of 11.8%. This high performance is due to the provision of sufficient and reliable features to the discriminator.

Table 5. Human pose detection on the MPII dataset based on PCKh with different AMIL network setting

| Methods | Head | Sho. | Elb. | Wri. | Hip | Knee | Ank. | Mean |
|---|---|---|---|---|---|---|---|---|
| AMIL with 1 FC | 95.8 | 90.3 | 80.5 | 74.3 | 77.6 | 69.7 | 62.8 | 79.6 |
| AMIL with 2 FC | 96.1 | 91.9 | 83.9 | 77.8 | 80.9 | 72.3 | 64.8 | 82.0 |
| AMIL with 3 FC | 98.1 | 96.0 | 90.1 | 88.1 | 91.1 | 88.3 | 84.9 | 91.8 |
| AMIL with 3 FC and adjust pooling | **98.8** | 96.5 | **92.5** | 88.5 | **91.5** | **88.8** | **85.8** | **92.3** |

Table 5 presents the performance of AMIL while having a varying number of FC layers. As the results show, the performance acquired with three FC layers and adjusted pooling is the best. In the proposed model, the adjust pooling has a key role in extracting deep features.

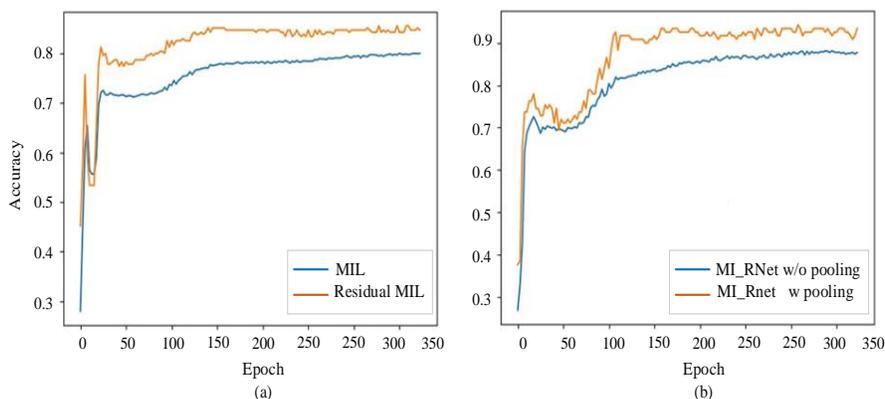

Fig. 6. PCKh on the MPII dataset. (a) Performance comparison of standard MIL and residual MIL. (b) Shows the performance of MI_RNet with and without the proposed adjust pooling.

### 4.3.2 Adversarial Training Performance

Figure 7 shows the confusion matrix comparison of the standard MIL and the proposed model on the LSP dataset. To check the advantage of the adversarial training, we compare the performance of AMIL with and without adversarial training on the MPII dataset. Figure 8(a) shows the significant improvement of the proposed model while using adversarial training and Figure 8(b) presents the effect of adversarial loss. As the results show, the AMIL has faster convergence and more stable performance while taking adversarial training. We also figure out that the learning rate decay approach is supportive; which resulted in more stable performance. The performance of AMIL on LSP and MPII datasets with and without the learning rate decay presented in Figure 8 (c) and (d) respectively.

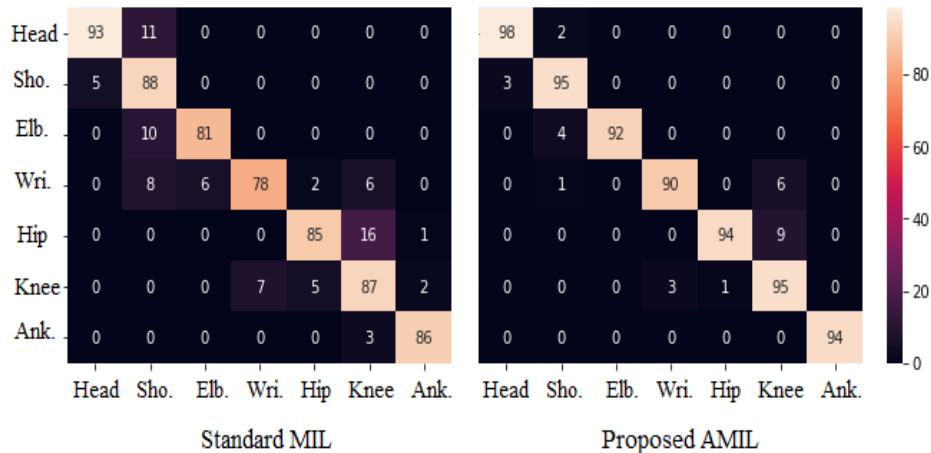

Fig. 7. Confusion matrix comparison on standard MIL and proposed AMIL on LSP dataset

Figure 9 presents the interchange concerning the computation cost and number of iterations of six pose estimation models on the MPII and LSP dataset. As the results show AMIL requires less computation costs as compared to the other state-of-the-art models. Between the iterations 200 and 250, Tompson et al. [12] performs similar to AMIL. However, after this period, AMIL outperforms the other models.

We believe that the presented algorithm is run by epochs. In respective epoch, we randomly partition the vertex set of particular image $V$ as $I$ mini-batches $V_1, V_2, \ldots , V_I$ , and in the $i$-th iteration, we run a forward pass to estimate the human pose for nodes in $V_i$ , a back propagation is used to compute the gradients, and update the history. In every epoch, the scanning of all the nodes is executed, rather than just training nodes, to check that the history of each node is updated at least once per epoch.

From the results, we notice that AMIL requires approximately $3\times$ fewer parameters and time to achieve comparable accuracy to the original MIL. Furthermore, AMIL did not use the depth-wise divisible FC, and just used the simple FC layers. It is possible to use AMIL as a meta-architecture to even obtain a more efficient network. Figure 10 shows the performance of proposed AMIL model in more complex cases. As the results show, the proposed model perfectly estimated the human poses in a single or group activities.

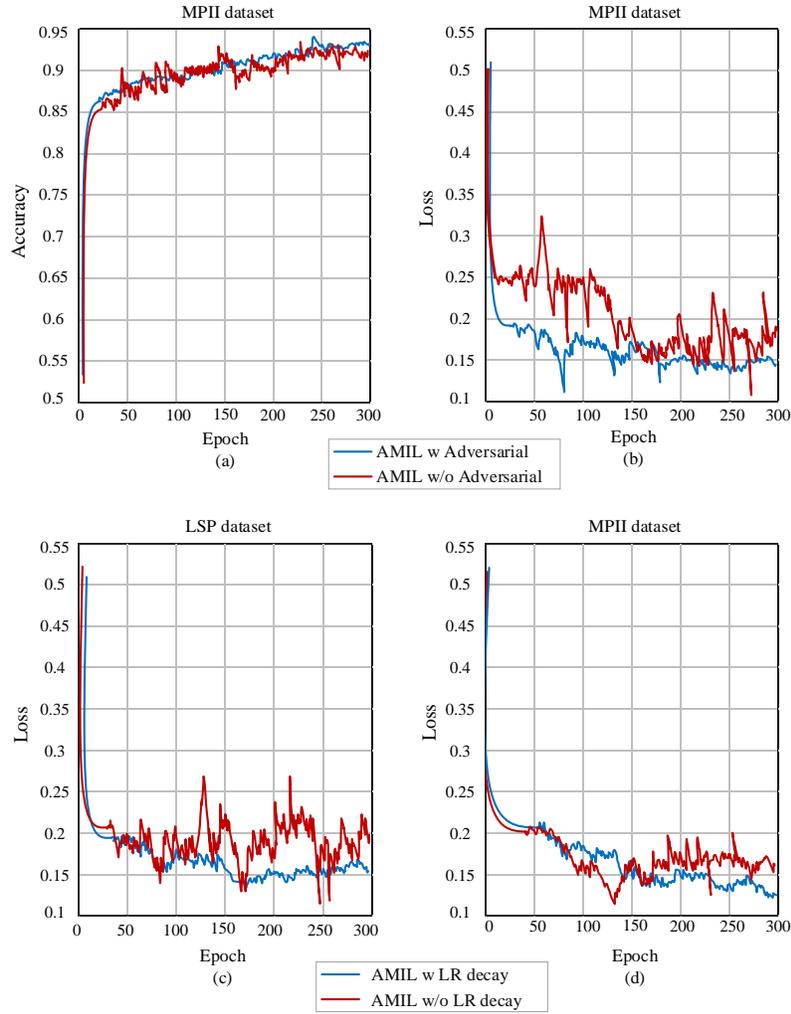

Fig. 8 (a) and (b). PCKh on the MPII dataset. Show the performance of AMIL with and without adversarial Training and adversarial Loss. (c) and (d) present the performance of AMIL with and without learning rate (LR) decay on LSP and MPII datasets.

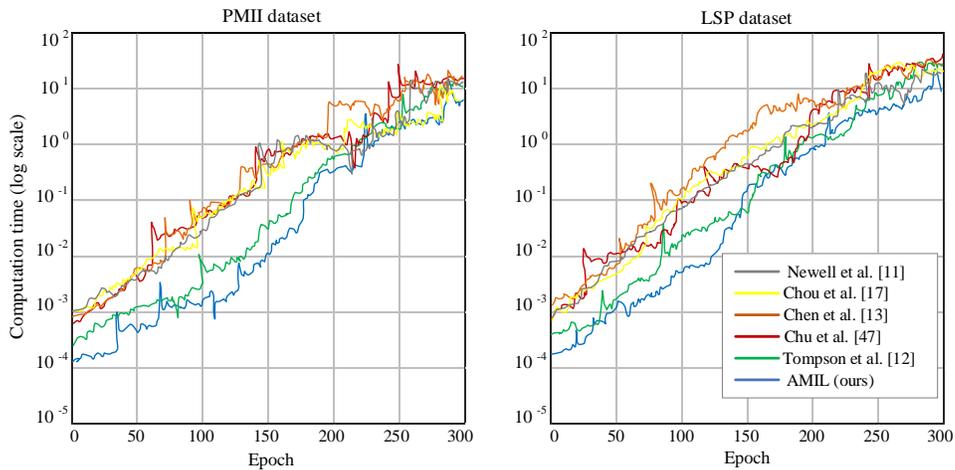

Fig. 9. Computation time vs number of iterations while training on MPII and LSP dataset

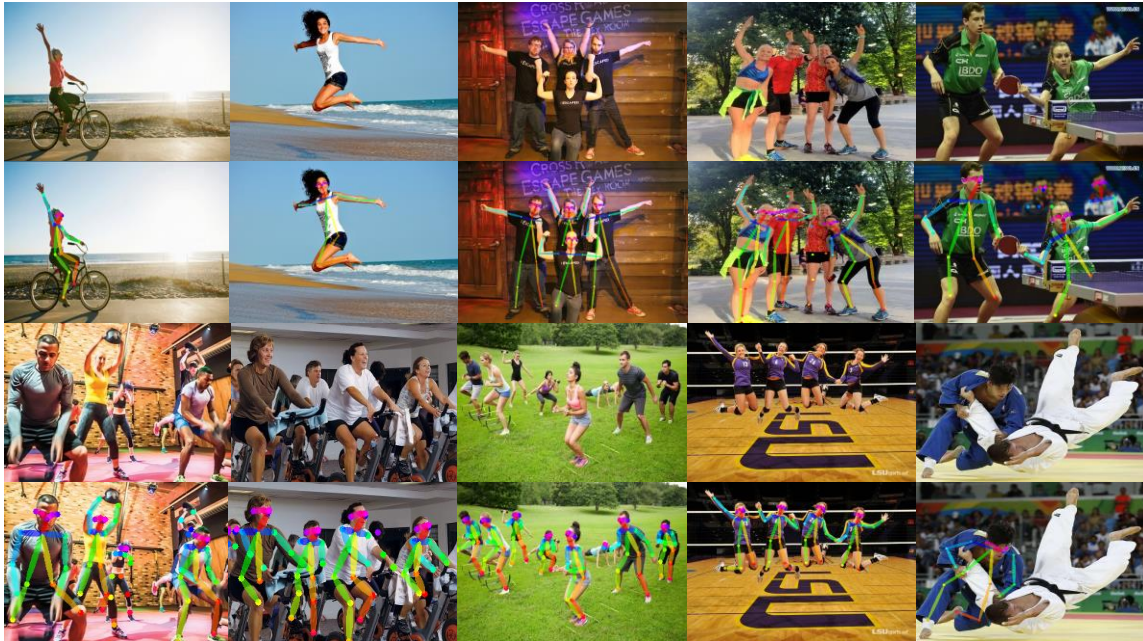

Fig. 10. Performance of AMIL in more complex cases. The blue lines indicate the right arm and green lines indicate the body right side, the Caribbean green lines indicate the left arm and yellow/orange lines indicate the left side of the body.

## 5 CONCLUSION

This work has presented an adversarial multi-instance neural network with adjust pooling to solve the human pose estimation problem. The proposed model has the combination of a generator and a discriminator with a similar architecture. The generator is operates based on the predicted feature heatmaps of the human body keypoints, and the discriminator is to distinguish implausible poses and advice useful hints to the generator for improving the heatmaps. After completion of the training, we can remove the discriminator, hence it does not affect the time of other tasks. We evaluated AMIL on two widely used human pose estimation benchmark datasets and the overall results proved that the proposed model outperformed several state-of-the-art approaches and generated better human pose prediction. We will explore hierarchical learning models in the future to incorporate the structural information into the deep models. Additionally, we plan to apply the proposed model to more extensive real applications, such as image segmentation and weakly supervised learning.